\documentclass[sigconf,number]{acmart}
\setcopyright{none}
\renewcommand\footnotetextcopyrightpermission[1]{}
\usepackage{enumitem}
\usepackage{tikz}
\usepackage{pgf-pie}
\usepackage{subcaption}
\usepackage{algorithm}
\usepackage{algpseudocode}

\usepackage{amsmath}
\usepackage{mathtools}
\usepackage{amsthm}
\usepackage{fixmath}
\usepackage{amsmath}
\usepackage{amsfonts}
\usepackage{pifont} 
\usepackage{xcolor} 
\usepackage{mathtools}

\usepackage{multicol}
\usepackage{multirow}

\AtBeginDocument{%
  \providecommand\BibTeX{{%
    \normalfont B\kern-0.5em{\scshape i\kern-0.25em b}\kern-0.8em\TeX}}}

\begin{document}
\pagestyle{plain}

\title{CHOSEN: Compilation to Hardware Optimization Stack for Efficient Vision Transformer Inference}







\author{Mohammad Erfan Sadeghi}
\authornote{These authors contributed equally to this research.}

\affiliation{%
  \institution{University of Southern California}
  \city{Los Angeles}
  \state{California}
  \country{USA}
  \postcode{90089}
}
\email{sadeghim@usc.edu}

\author{Suhas Somashekar}
\authornotemark[1]

\affiliation{%
  \institution{University of Southern California}
  \city{Los Angeles}
  \state{California}
  \country{USA}
  \postcode{90089}
}
\email{ssomashe@usc.edu}

\author{Arash Fayyazi}
\authornotemark[1]

\affiliation{%
  \institution{University of Southern California}
  \city{Los Angeles}
  \state{California}
  \country{USA}
  \postcode{90089}
}
\email{fayyazi@usc.edu}

\author{Armin Abdollahi}
\affiliation{%
  \institution{University of Southern California}
  \city{Los Angeles}
  \state{California}
  \country{USA}
  \postcode{90089}
}
\email{arminabd@usc.edu}

\author{Massoud Pedram}
\affiliation{%
  \institution{University of Southern California}
  \city{Los Angeles}
  \state{California}
  \country{USA}
  \postcode{90089}
}
\email{pedram@usc.edu}








\begin{abstract}
Vision Transformers (ViTs) represent a groundbreaking shift in machine learning approaches to computer vision. Unlike traditional approaches, ViTs employ the self-attention mechanism, which has been widely used in natural language processing, to analyze image patches. Despite their advantages in modeling visual tasks, deploying ViTs on hardware platforms, notably Field-Programmable Gate Arrays (FPGAs), introduces considerable challenges. These challenges stem primarily from the non-linear calculations and high computational and memory demands of ViTs. This paper introduces CHOSEN, a software-hardware co-design framework to address these challenges and offer an automated framework for ViT deployment on the FPGAs in order to maximize performance. Our framework is built upon three fundamental contributions: multi-kernel design to maximize the bandwidth, mainly targeting benefits of multi DDR memory banks, approximate non-linear functions that exhibit minimal accuracy degradation, and efficient use of available logic blocks on the FPGA, and efficient compiler to maximize the performance and memory-efficiency of the computing kernels by presenting a novel algorithm for design space exploration to find optimal hardware configuration that achieves optimal throughput and latency. Compared to the state-of-the-art ViT accelerators, CHOSEN achieves a 1.5x and 1.42x improvement in the throughput on the DeiT-S and DeiT-B models.

\end{abstract}

\maketitle

\section{Introduction}

The landscape of computer vision has been fundamentally transformed with the advent of deep learning, among which Vision Transformers (ViTs) \cite{DBLP:conf/iclr/DosovitskiyB0WZ21, DBLP:conf/icml/TouvronCDMSJ21, fayyazi2024neuroblend, DBLP:conf/iccv/LiuL00W0LG21} have emerged as a particularly promising approach. Unlike traditional convolutional neural networks (CNNs) that rely on local receptive fields, ViTs leverage the power of self-attention mechanisms to capture global dependencies within an image, enabling a more comprehensive understanding of visual data. This capability has placed ViTs at the forefront of research, demonstrating state-of-the-art performance across a wide range of tasks in computer vision. Overall, deep learning has revolutionized various domains by providing robust algorithms capable of learning complex patterns from large datasets, thus enabling unprecedented advancements in the application of artificial intelligence across numerous fields, from healthcare \cite{DBLP:journals/tim/BaraeinejadSVRH22} to recommendation systems \cite{zarch2024cadc} to scientific research.

ViTs rely on a series of identical encoder blocks to progressively extract complex features from an image. These encoder blocks consist of two principal components: Multi-headed Attention (MHA) and Feed-Forward Network (FFN), each prefaced with a layer normalization block. Embedded within MHA and FFN are linear layers, GELU, and softmax, integrated via two residual connections that bookend the normalization stages. The output of the final encoder block goes through a classifier to obtain the class predictions.

Implementing Vision Transformers (ViTs) and other transformer-based models on Field-Programmable Gate Arrays (FPGAs) presents unique challenges that stem from the architectures' complexity and resource demands. The primary hurdles include the high demand for memory due to the extensive number of parameters and the intensive computation required for processing the self-attention across image patches. FPGAs, although flexible and capable of parallel processing, often struggle with limited on-chip memory and bandwidth, which can bottleneck the performance of ViTs. Additionally, the static nature of FPGA architectures complicates the implementation of the dynamically varying computational patterns of ViTs.

A wide range of methods has been explored to improve the efficiency of ViTs, including approaches like quantization \cite{DBLP:conf/nips/LiuWHZMG21}, model pruning \cite{DBLP:conf/aaai/0004HWCCC22}, token pruning \cite{heo2024trainingfreeaccelerationvitsdelayed}, and low-rank approximation \cite{azizi2024memoryefficientvisiontransformersactivationaware}. However, deploying ViTs in practical applications requires innovative approaches in hardware optimization, such as strategic memory management and multi-kernel design for increased throughput. The CHOSEN framework integrates these strategies effectively in addition to its compiler to offer a complete stack for automating the FPGA deployment of ViTs. The main contributions of this paper are summarized below.
\begin{itemize}[leftmargin=10pt]
        \vspace{-1.5mm}
    \item We present an end-to-end framework called CHOSEN, which generates high-performance ViT accelerators suited to a target FPGA device from a high-level description of the ViT model in any machine learning framework, such as PyTorch while making effective use of the available FPGA resources and memory bandwidth. Note that CHOSEN can handle any transformer-based model, while this paper focuses on ViTs.
   \item We develop optimized synthesizable C++ templates, achieving high-throughput accelerator designs on FPGAs by focusing on multi-kernel design to maximize bandwidth utilization and keep all the used resources busy. The CHOSEN ViT follows static scheduling and has a different memory layout for each operation to achieve full burst read data from off-chip memory banks.  
   \item We introduce a powerful compiler (CHOSEN Compiler) for mapping a given transformer-based model running on any dataset onto our optimized accelerator design by converting the network model to a computational graph, scheduling the graph's execution, and optimizing its nodes by leveraging combining matrices if we have enough resources. 
    \item Compared to the state-of-the-art work, using our optimizations focused on multi-kernel design and maximizing bandwidth utilization, we achieve significant improvement in the throughput.
\end{itemize}

\section{Related Work}
\label{sec:rel_work}
This section includes prior works on ViT architectures, compiler optimizations, and approximations of non-linear functions.
\subsection{Compiler}
\label{rel_compiler}
Previous efforts have explored software-hardware co-design frameworks for efficiently deploying ViTs. The VAQF framework \cite{DBLP:journals/corr/abs-2201-06618} dynamically adjusts weights precision, activations precision, and hardware settings based on FPGA resources and target FPS. ViTCoD \cite{DBLP:conf/hpca/YouSSYZZLLL23} proposes methods for pruning and reconfiguring the attention maps in ViTs to create either denser or sparser patterns, which helps in managing computational workloads more effectively. However, both VAQF \cite{DBLP:journals/corr/abs-2201-06618} and ViTCoD \cite{DBLP:conf/hpca/YouSSYZZLLL23} do not leverage off-chip (DDR) memory parallelism and do not introduce a set of approximations for efficiently calculating non-linear functions on FPGAs. 

\subsection{Hardware architecture}
\label{rel_hardware}
Several architectures have been proposed to accelerate Vision Transformer (ViT) and Transformer inference on FPGAs, typically employing quantization techniques that reduce the model size and computational requirements by converting weights and activations to 8-bit representations such as Auto-ViT-Acc presented in \cite{DBLP:conf/fpl/LiSLMYX0LLWLF22}.
ME-ViT \cite{DBLP:conf/hipc/MarinoZP23} is a state-of-the-art ViT accelerator that mitigates the high-bandwidth demands of Vision Transformers (ViTs) during inference by employing a single-load policy and utilizes multi-purpose buffers within a memory-efficient processing element (ME-PE) to minimize memory traffic for ViT accelerators on FPGAs. They achieve this by avoiding storing and reloading in matrix multiplications and buffering the intermediate results. However, ME-ViT does not explore multiple DDR banks to achieve higher bandwidth.

\section{CHOSEN Framework}
\label{sec:framework}
CHOSEN is a compilation framework that enables the optimization and deployment of ViTs on FPGAs. It takes a high-level description of the ViT model specified in PyTorch and generates a high-performance ViT accelerator design tailored to the FPGA device. The CHOSEN compiler optimizes the inference graph associated with the ViT model, aligning it with the accelerator design. Next, it produces hardware parameters for the accelerator and a static schedule for executing its various operations. A detailed explanation of the state space exploration process used to find the optimal hardware parameters is provided in Section \ref{sec:compiler}. Finally, using optimized accelerator component templates, CHOSEN compiler generates synthesizable C-level code descriptions that can be used to generate an FPGA bitstream. 

\section{Accelerator Design Optimizations}
\label{sec:HW}
This section focuses on improving the efficiency and performance of a ViT accelerator using FPGA device-aware optimizations. Solutions include data placement optimizations to reduce the number of accesses to DRAM, the use of the on-chip memory to balance computation vs. memory bandwidth, and approximations for nonlinearities functions.
\subsection{Kernel Architecture}
\label{subsec:ker_arch}
\begin{figure}[h]
\centering
\vspace{-3mm}
\includegraphics[width=0.8\linewidth]{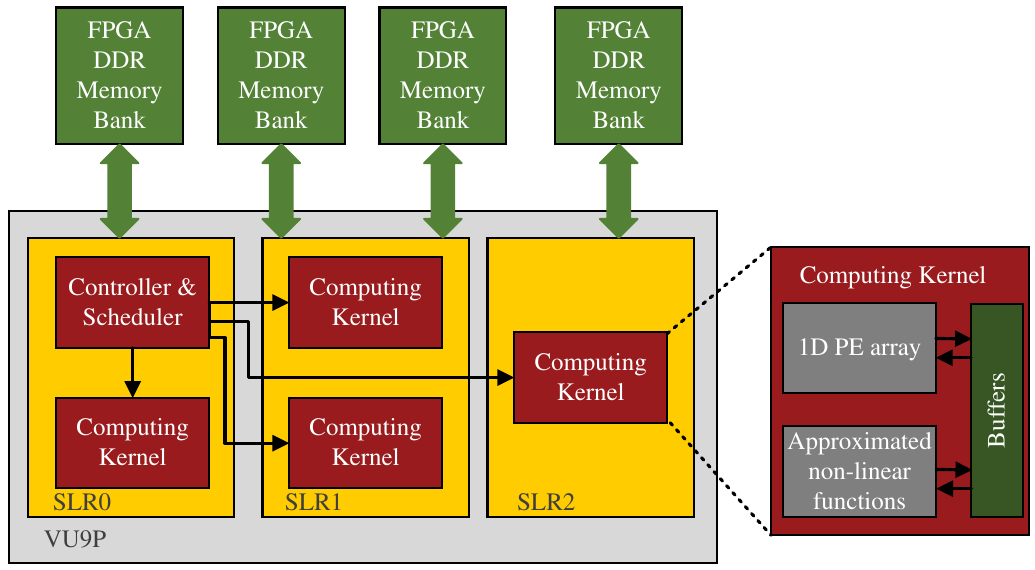}
\vspace{-3mm}
\caption{Overview of Multi-kernel CHOSEN-ViT architecture on the VU9P FPGA.}
\label{fig:arch}
\vspace{-3mm}
\end{figure}

Our accelerator is a multi-kernel design where each kernel performs the same operation. Each kernel comprises (i) a 1D array of processing elements (PEs), which
are responsible for executing the MAC operations associated with the matrix multiplications, (ii) a
memory hierarchy feeding data to the said 1D array compromising of register files, the on-chip memory (Block RAMs and Ultra RAMs on FPGA devices), and external off-chip memory (DRAM), and (iii) a processing unit (PU) that performs approximated non-linearity and other required operations like the addition of skip and main paths as shown in Fig.~\ref{fig:arch}

For our 1D array of PEs, we adopt the architecture proposed in \cite{DBLP:conf/fpga/LichtKH20}. Their architecture is used to perform efficient multiplication of \( \mathbf{A}\) and \( \mathbf{B}\) matrices where $\mathbf{A} \in \mathbb{R}^{n \times k}$ and $\mathbf{B} \in \mathbb{R}^{k \times m}$. The output matrix is $\mathbf{C} \in \mathbb{R}^{n \times m}$. The computational resources are organized into \( P_n\) of 1D processing elements, which encapsulate a vector operation of \( P_m\) compute elements (i.e., DSPs on the FPGA). To support a hierarchical hardware design, each matrix is further decomposed into several levels of tiling. 
We tile matrix \( \mathbf{A}\) and \( \mathbf{B}\)  column and row-wise with factor of \( T_n\) and \( T_m\), respectively.  

\subsection{Non-linear approximations}
\label{rel_nonLinear}
Implementing the non-linear functions of the Vision Transformer (ViT) presents significant challenges due to their inherent computational complexity. To enable an efficient FPGA-based implementation of ViT, we employed a set of approximations for non-linearities as presented in PEANO-ViT \cite{sadeghi2024peano}. Specifically, they approximated the inverse square root function in layer normalization using bit manipulation operations and pre-stored fractional powers of two. For the softmax function, they utilized a Padé-based approximation of the exponential function, complemented by bit manipulation techniques to eliminate division operations. Additionally, the GELU activation function was replaced with a piece-wise linear approximation that closely mimics the original function. These strategic approximations not only facilitate the efficient implementation of ViTs on FPGAs but also exhibit negligible degradation in the model’s accuracy.
We used all of these approximations in the CHOSEN ViT implementation to achieve high throughput and balance usage of DSP and LUTs in our design.

\subsection{Memory Layout and Data Placement}
\label{sub:MAC-data-placement}
To utilize the off-chip memory bandwidth, we group activation data (\( \mathbf{A}\)) as well as weights (\( \mathbf{B}\)) before sending them to the on-chip memory. For both matrices (e.g., \( \mathbf{A}\) and \( \mathbf{B}\)), we group \( \left\lfloor \frac{AXI\_WIDTH}{2\times DW} \right\rfloor \) of them in the column dimension. By this packing, we perform full burst read/write of data from/to the memory banks and utilize the maximum possible burst size (512-bit width) allowable on the Advanced eXtensible Interface (AXI) bus of the target FPGA board. 

\begin{figure}[tbp]
     \centering
     \vspace{-2mm}
     \begin{subfigure}[b]{0.38\textwidth}
         \centering
         \includegraphics[width=\textwidth]{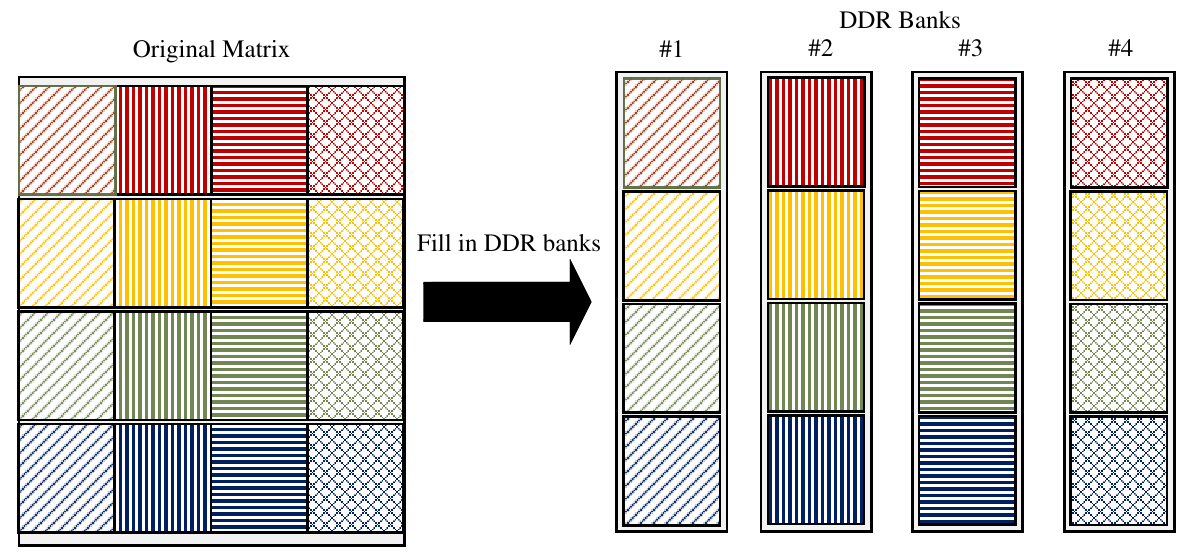}
         \caption{Storing a matrix in four DDR banks.}
         \vspace{-2mm}
         \label{fig:ddr}
     \end{subfigure}
     \hfill
     \begin{subfigure}[b]{0.37\textwidth}
         \centering
         \includegraphics[width=\textwidth]{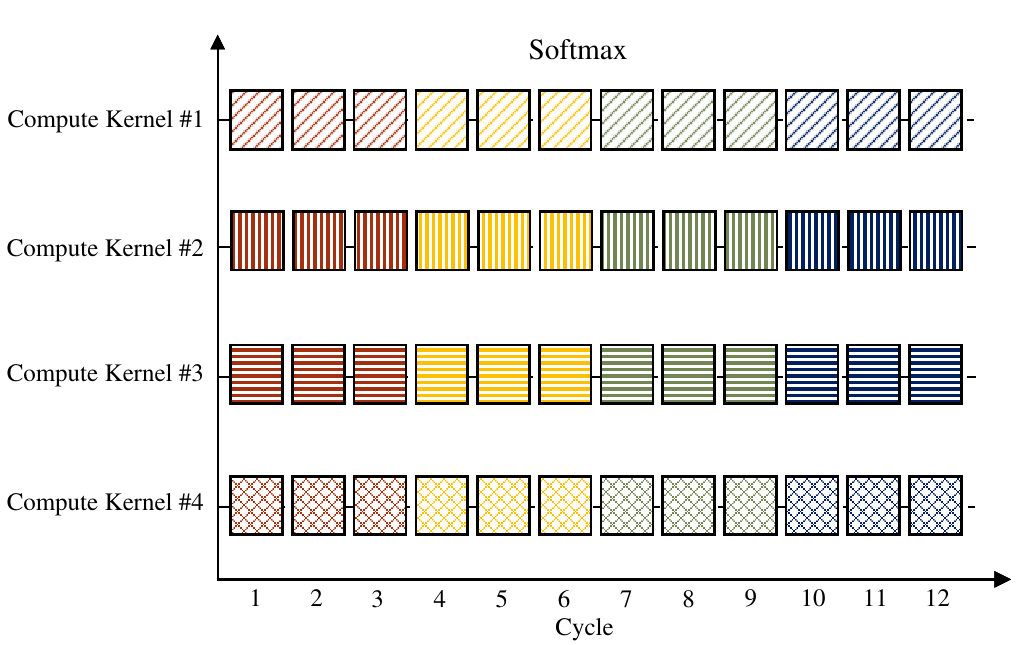}
         \caption{Scheduling for Softmax operations.}
         \vspace{-2mm}
         \label{fig:sc_soft}
     \end{subfigure}
     \hfill
     \begin{subfigure}[b]{0.30\textwidth}
         \centering
         \includegraphics[width=\textwidth]{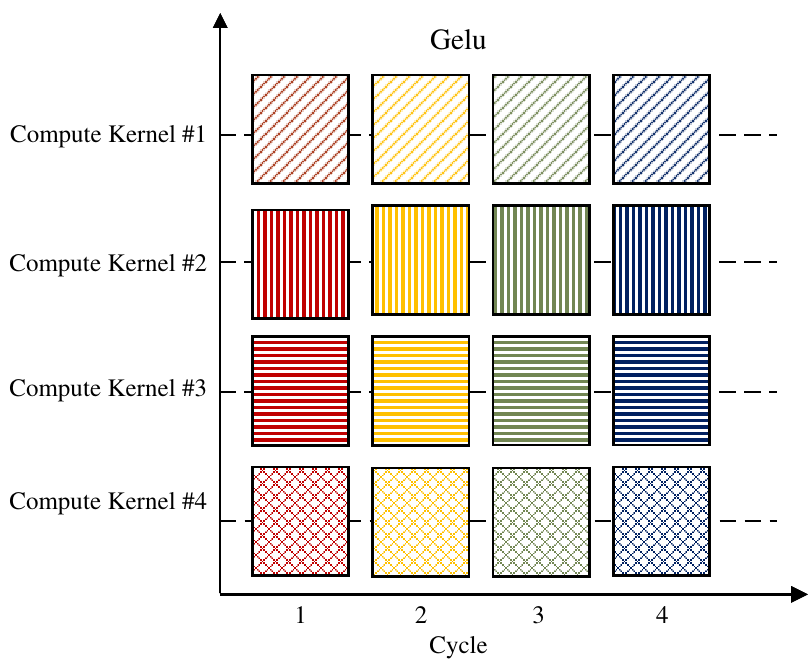}
         \caption{Scheduling for LayerNorm operations.}
         \vspace{-3mm}
         \label{fig:sc_gelu}
     \end{subfigure}
     \hfill
     \begin{subfigure}[b]{0.30\textwidth}
         \centering
         \includegraphics[width=\textwidth]{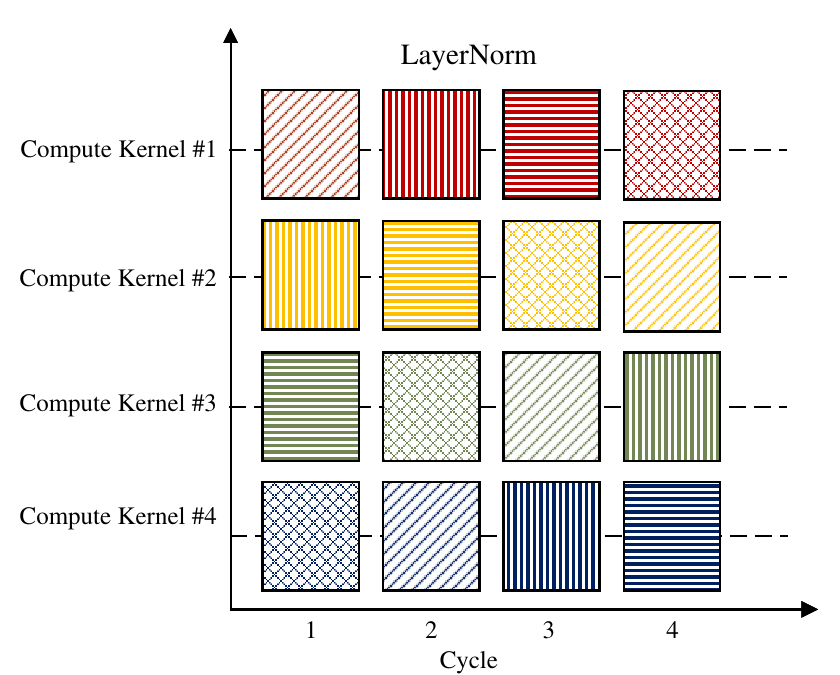}
         \caption{Scheduling for Gelu operations.}
         \vspace{-1mm}
         \label{fig:sc_ln}
     \end{subfigure}
     \vspace{-3mm}
        \caption{\small Scheduling of different operations while using multiple DDR banks. Different patterns represent different DDR memory banks, while different colors represent different rows.}
        \label{fig:scheduling}
        \vspace{-5mm}
\end{figure}

We use all available DDR banks to load and store the data to maximize the bandwidth. We divide the original matrix column-wise into $BN$, which denotes the total number of available DDR banks and is 4 in our case (see Section \ref{sec:res} for more details regarding the target FPGA board), smaller matrices and store each smaller matrix in one DDR bank as shown in Fig.~\ref{fig:ddr}. In the provided example shown in Fig.~\ref{fig:scheduling}, we have 4 DDR banks and 4 hardware kernels. For operations outside of the head, like key, query, value matrices, and Gelu computations, we bring the data from the same row but with different DDR banks and pass them to different hardware kernels for computation (see Fig.~\ref{fig:sc_gelu}).
For the head-wise operations, including softmax, we bring the data from the same bank and the same row \( \left\lceil \frac{N_h}{BN} \right\rceil \) times to finish the required computations for one row where \(N_h\) represents the number of heads. For instance, in the case of ViT-B, it takes three rounds to finish a row, as shown in Fig.~\ref{fig:sc_soft}. 

With respect to layerNorm, where each hardware kernel requires the whole row to complete its computation, such as calculating the mean and variance, CHOSEN offers efficient rotating scheduling (see Fig.~\ref{fig:sc_ln}). Hardware kernel \#1 operates on the first part of row \#1 from DDR bank \#1, while hardware kernel \#2 performs on the second part of row \#2 from DDR bank \#2. Similarly, other kernels access corresponding data. In the next round, Hardware kernel \#1 operates on the second part of row \#1 from DDR bank\#2, while hardware kernel \#2 performs on the third part of row \#3 from DDR bank \#3. In this approach, we maximize the utilization of potential bandwidth and keep all hardware kernels busy. 

\section{CHOSEN Compiler}
\label{sec:compiler}

This section elaborates on our compiler framework, designed to enhance the execution efficiency of transformer models on FPGAs. The compiler adeptly utilizes the underlying FPGA architecture, focusing on optimal scheduling, precise operation mapping to hardware resources, and minimizing the data movement, thereby facilitating high-performance Vit acceleration hardware on FPGAs.

\subsection{Compiler Design and Execution Framework}

Our compiler framework starts with a high-level description of a ViT model and converts it into a hardware-accelerated implementation tailored to the target FPGA's architecture. The transformation of a ViT model into an FPGA-compatible format unfolds through several steps, as explained below.
\begin{enumerate}[leftmargin=10pt]
    \item \textbf{Model Conversion:} Initially, the transformer model, developed using frameworks such as PyTorch or TensorFlow is transformed into a Directed Acyclic Graph (DAG). This graph outlines the model’s computational dependencies and data flow, providing a basis for further analysis and optimization.
    
    \item \textbf{Graph Analysis and Optimization:} The compiler conducts an in-depth analysis of the DAG to classify operations and deduce essential characteristics such as operation types, data sizes, and dependency chains. This information is crucial for optimizing the execution schedule and effectively mapping the operations onto the FPGA. It is worth mentioning that the CHOSEN Compiler can optimize its nodes by leveraging combining matrices if we have enough resources. For instance, we can calculate $k$ matrix in one shot and then separate it column-wise for each head or calculate k head-wise and name them $k\_h$ matrices. If we have enough resources on the target FPGA, we can even concatenate the weight matrices of $k$, $q$, and $v$ to increase the second dimension of the second matrix to achieve potentially higher $T_m$. In parallel, we can increase the batch size to achieve potentially higher $T_n$. All of these cases result in larger design space.
    
    \item \textbf{Design Space Exploration:} The compiler assesses various hardware configurations and operational parameters using a custom heuristic-based algorithm for design space exploration. This phase ascertains the optimal tiling and parallelization strategies, ensuring they align with both the hardware's capabilities and the computational demands of the model.

    
    \item \textbf{Code Generation and Deployment:} The final phase is the generation of synthesizable C++ templates and hardware description code, culminating in the production of the FPGA bitstream. This transformation converts the high-level software model into a practical hardware implementation.
\end{enumerate}


    

\subsection{Compiler Optimization for Matrix Multiplications}
Optimizing matrix multiplication for hardware implementations necessitates a fine-grained analysis of computational resources and data management. This section describes our approach to determining optimal tile parameters, which is crucial for enhancing the performance of matrix multiplication computations, especially in transformer models. We present a new algorithm that balances the computational load across processing elements while minimizing memory access latencies and maximizing throughput.

\begin{algorithm}[tb]
\caption{Exhaustive Search for Finding Optimal Tile Parameters}
\label{alg:optimal_tile_size_base}
\begin{algorithmic}[1]
\State \textbf{Input:} Model Graph \( G \), Data Width \( DW \), Memory Constraints \( M \)
\State \textbf{Output:} Optimal tile sizes \( T_n, T_m \), Parallelism Factors \( P_n, P_m \)
\State Initialize minimum cost: \( \text{min\_cost} \leftarrow \infty \)
\State Calculate \( P_m \): \( P_m \leftarrow \left\lfloor \frac{AXI\_WIDTH}{2 \times DW} \right\rfloor \) \Comment{Fixed value based on data width}
\State Set \text{optimal\_pm} \( \leftarrow {P_m}\)
\For{\( T_m \) \(\in\) \(\left\{ \text{feasible } T_m \text{ sizes} \right\}\)}
\For{\( P_n \) \(\in\) \(\left\{ \text{feasible } P_n \text{ sizes} \right\} \) where \( P_n < \frac{T_m}{P_m} \)}

        \For{\( T_n \) \(\in\) \(\left\{\text{feasible } T_n \text{ sizes}\right\}\)}
            \State Cost:\text{ cost} \( \leftarrow \text{compute\_cost}(P_n, P_m, T_n, T_m, G, M) \)
            \If{\( \text{cost} < \text{min\_cost} \)}
                \State Update minimum cost and optimal sizes:
                \State \( \text{min\_cost} \leftarrow \text{cost} \)
                \State \( \text{optimal\_pn} \leftarrow P_n \)
                \State \( \text{optimal\_tn} \leftarrow T_n \)
                \State \( \text{optimal\_tm} \leftarrow T_m \)
            \EndIf
        \EndFor
    \EndFor
\EndFor
\State \Return \( \text{optimal\_pn},\text{optimal\_pm} ,\text{optimal\_tn}, \text{optimal\_tm} \)
\end{algorithmic}
\end{algorithm}

 The algorithm for computing optimal tile parameters, as presented in Algorithm~\ref{alg:optimal_tile_size_base}, outlines a systematic approach to identifying the most effective tiling and parallelization strategies. The algorithm explores a design space defined by the constraints of the available hardware resources, the properties of the matrix operation, and the compiler's high-level understanding of the transformer model represented as a Directed Acyclic Graph (DAG).



\subsubsection{Initialization and Fixed Parameters}

The parallelism factor \( P_m \) is computed as \( P_m = \left\lfloor \frac{AXI\_WIDTH}{2 \times DW} \right\rfloor \), establishing a fixed number of computation units based on the data and AXI width. For DDR4 memory, a minimum of 512 bits as AXI width must be transferred to make up for the I/O clock multiplier, and much longer bursts are required to saturate DDR bandwidth in practice.

\subsubsection{Exploration of Tiling Parameters}

The exploration of tile sizes \( T_n \) and \( T_m \) and the parallelism factor \( P_n \) is conducted within feasible ranges determined by the hardware specifications and the nature of the matrix operations. The nested loops in the algorithm reflect an exhaustive search within these ranges, ensuring that each combination of \( T_n \), \( T_m \), and \( P_n \) is evaluated for its performance:


\begin{enumerate}[leftmargin=10pt]
        \vspace{-1.5mm}
    \item \textbf{Tiling Factor \( T_m \)}: Represents the number of elements from one row of matrix \( B \) that are loaded into the compute units. Each feasible size is evaluated to find the optimal configuration of compute and memory resources.
    \item \textbf{Parallelism Factor \( P_n \)}: Reflects the number of processing elements and is constrained by \( P_n < \frac{T_m}{P_m} \), ensuring that data for the next round of computation is already distributed to all PEs.
    \item \textbf{Tiling Factor \( T_n \)}: Corresponds to the number of elements from one column of matrix \( A \) loaded per cycle, which, when combined with \( T_m \), should not exceed the total memory capacity \( S \) as \( T_n \times T_m \leq S \).
\end{enumerate}

\subsubsection{Cost Function for Optimizing Multiple Matrix Multiplications}

In optimizing transformer models, particularly for matrix multiplications within attention mechanisms, this function is designed to minimize latency by balancing the computational load across processing elements (PEs) and computation units within PEs. The cost function, \( \text{Latency} \), is calculated as follows:
\begin{equation}
\begin{aligned}
    &Latency = \frac{Total Cycles}{Frequency} \\
    &= \frac{T_n \times T_m \times k \times num Tiles Row \times num Tiles Col \times kernel Factor}{P_n \times P_m \times Frequency}
\end{aligned}
\end{equation}
Where \(Total Cycles\) is the total number of cycles required to perform all multiplication operations. \( Frequency \) is the operating frequency of the hardware accelerator.

The calculation of \( Total Cycles \) includes other parameters that are outlined in the following.
i) Computation of Total Operations: The total operations required for a matrix multiplication between matrices \( A \) and \( B \) are determined by the tile sizes \( T_n \) and \( T_m \), and the dimensions of the matrices involved. We modeled this calculation as:
$totalOps = opsPerTile \times numTilesRow \times numTilesCol$, where each tile's operations,   
$opsPerTile = T_n \times T_m \times k$,
where \( sharedDimension \) denoted as \( k \) is the inner dimension shared between matrices \( A \) and \( B \). \( numTilesRow \) (\( numTilesCol \)) define how many tiles matrix A (B) are divided into along its rows (columns). This equation reflects the multiplication operations needed to compute the product of submatrices defined by the tiles.

ii) Distribution of Operations Across Hardware Resources: The division of \( totalOps \) by \( P_n \) and \( P_m \) reflects the distribution of operations across multiple processing elements (PEs) and computation units (CU), respectively:
\[
adjusted\_cycles = \frac{totalOps}{P_n \times P_m} \times kernelFactor
\]
Where \( P_n \) (\( P_m \)) is the number of processing elements (computation units per processing element).

This division is critical because it ensures that the computational load is balanced across all available hardware resources, thereby optimizing the utilization of the FPGA's resources and minimizing the computation time.

iii)Adjustment for Multi-Kernel:
The \( kernelFactor \) incorporates the multi-kernel nature of designed architecture, particularly pertinent for devices like the Xilinx UltraScale+ FPGA. For transformer models, where matrix multiplication is performed within multiple attention heads, this parameter is calculated as follows:
\[
kernelFactor = \left\{
\begin{array}{ll}
\lceil \frac{\text{num\_heads}}{\text{num\_kernels}} \rceil & \text{if head\_flag is true} \\
\frac{1}{\text{num\_kernels}} & \text{otherwise}
\end{array}
\right.
\]
This adjustment is necessary to align the computational load with the physical distribution of computational resources across different kernels in the FPGA, ensuring efficient data distribution and parallel processing.





The exhaustive search approach to determining optimal tile parameters, as described in the Algorithm~\ref{alg:optimal_tile_size_base}, explores all potential combinations of \(T_n\), \(T_m\), \(P_n\), and \(P_m\) that satisfy hardware constraints. 
The number of valid configurations can be exceedingly high in practical scenarios. For instance, even for small models like ViT Tiny on Xilinx UltraScale+ VU9P FPGA, we have approximately 116,144 valid combinations of tiling parameters, and it can go up to 586,554 valid combinations for ViT Small on the same FPGA. Evaluating the cost for each of these combinations demands extensive computational resources and time, resulting in a significant overhead on the compiler.

\subsection{CHOSEN Algorithm for Efficient Design Space Exploration}
To address the inefficiencies of the exhaustive search, a novel heuristic-based search algorithm is proposed to reduce the number of evaluation points required to converge to an optimal solution. Algorithm~\ref{alg:optimal_tile_size} is specifically tailored to optimize tiling parameters for deploying Vision Transformers (ViTs) on FPGA platforms, addressing constraints such as limited memory and parallel processing. The algorithm starts with a set of randomly generated tiling parameters that fit within the FPGA's specifications. It then iteratively refines these parameters to minimize latency for ViT inference.

The key components of our implementation are:
i)\textbf{Initial Param Set Creation}: Parameters are generated within feasible ranges specific to FPGA constraints, ensuring a diverse starting point. ii)\textbf{Performance Evaluation}: Each configuration's effectiveness is measured specifically by its impact on latency reduction in ViT applications. iii)\textbf{Evaluation Cache}: Stores latency results of previously tested configurations to avoid redundant computations, optimizing the refinement process. iv)\textbf{Refinement and Preservation}: Ensures configurations yielding the best latency results are carried forward. v)\textbf{Adaptive Refinement Strategy}: Prioritizes adjustments in parameters such as \(P_n\)  and \(T_m\) , which are empirically linked to better performance in the current application, guiding the set towards potentially optimal regions of the design space.

Our approach significantly reduces the number of evaluations required. For instance, in the case of ViT Tiny, the proposed algorithm converged to the optimal configuration with approximately 2,000 evaluations, while for ViT Small, around 3,700 evaluations were necessary. This represents a drastic reduction compared to the potential millions of evaluations required by an exhaustive search, demonstrating our algorithm's efficiency in finding near-optimal solutions within a much smaller computational budget.

\begin{algorithm}[tb]
\caption{CHOSEN's Algorithm for Finding Optimal Tile Parameters}
\label{alg:optimal_tile_size}
\begin{algorithmic}[1]
\State \textbf{Input:} ViT Model Graph \( G \), Data Width \( DW \), FPGA Memory Constraints \( M \)
\State \textbf{Output:} Optimal tile sizes \( T_n, T_m \), Parallelism Factors \( P_n, P_m \)
\State Initialize set size: \( \text{set\_size} \)
\State Number of iterations: \( \text{iterations} \)
\State Preservation size: \( \text{preservation\_size} \)
\State Calculate \( P_m \): \( P_m \leftarrow \left\lfloor \frac{AXI\_WIDTH}{2 \times DW} \right\rfloor \) \Comment{Fixed value}
\State Extract specific layer information from \( G \)
\State Initialize set with random configurations tailored to FPGA specs
\For{\( \text{iteration} = 1 \) \textbf{to} \( \text{iterations} \)}
    \State Evaluate performance of each configuration in terms of latency
    \State Preserve high-performance configurations based on latency improvements
    \State Generate new set by adaptive strategy 
    \State Replace old set with the new one
\EndFor
\State Best configuration \( \leftarrow \) configuration from the final set with the lowest latency
\State \Return \( \text{optimal\_pn},\text{optimal\_pm},\text{optimal\_tn},\text{optimal\_tm} \)
\end{algorithmic}
\end{algorithm}

\section{Results and Discussions}
\label{sec:res}
In this study, CHOSEN-ViT's hardware model was implemented on a Xilinx UltraScale+ VU9P board running at 200 MHz. This FPGA device includes 64 GiB DDR4 ECC-protected memory with a dedicated PCIe x16 connection. There are four DDR banks. To evaluate the performance of CHOSEN-ViT, we employed the publicly available ImageNet-1K dataset \cite{DBLP:conf/cvpr/DengDSLL009} and two different model architectures, namely DeiT \cite{DBLP:conf/icml/TouvronCDMSJ21} and ViT \cite{DBLP:conf/iccv/LiuL00W0LG21}, across various sizes (small, base, and large). It is important to point out that our experimental setup does not require extensive retraining for the proposed approximations.

\subsection{Comparison of CHOSEN's Algorithm to the Exhaustive Search}

\begin{figure}[tb]
\centering
\includegraphics[width=0.8\linewidth]{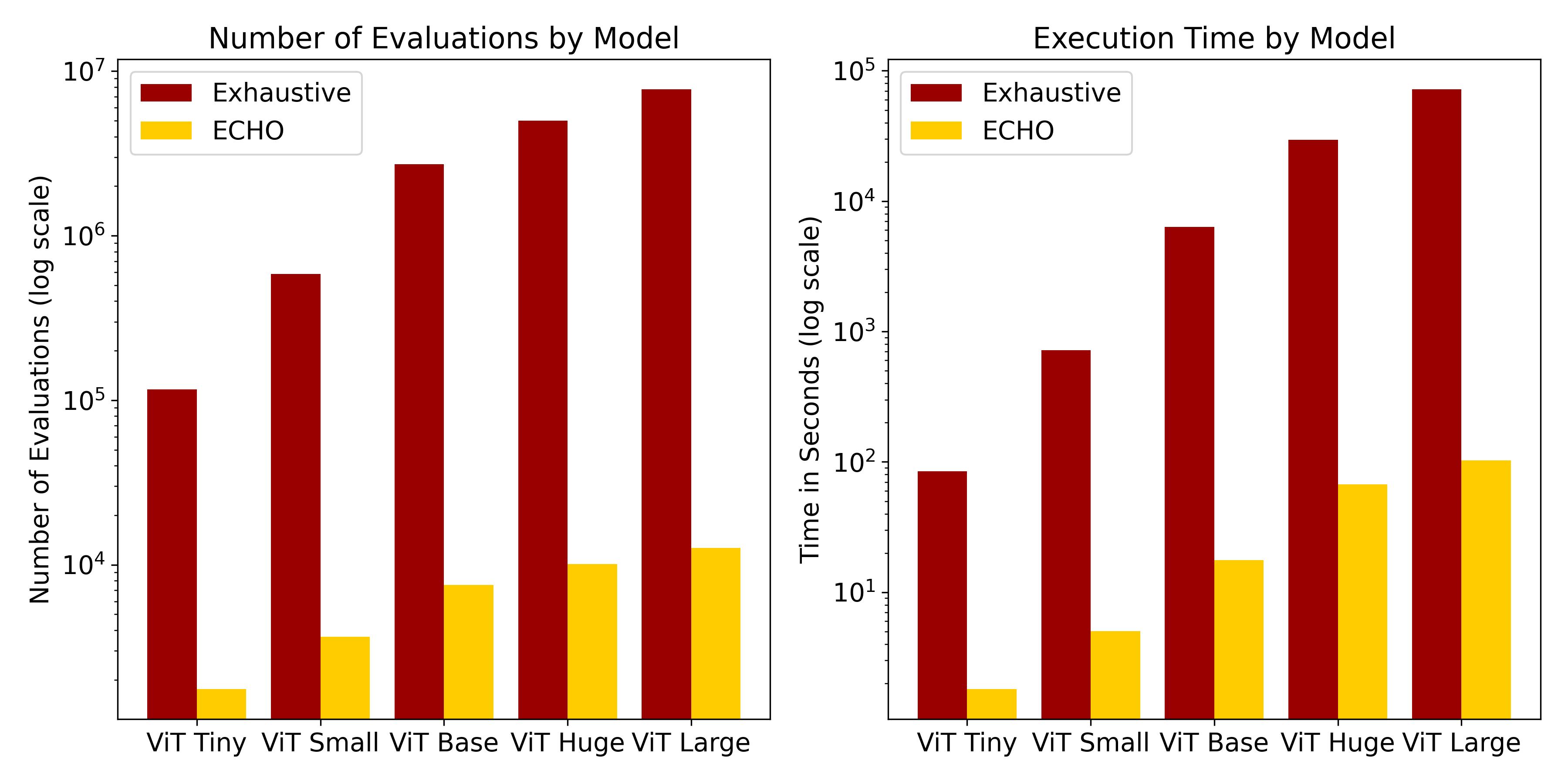}
\vspace{-4mm}
\caption{Efficiency comparison between exhaustive and CHOSEN search methods across different ViT models. }
\label{fig:comparison_chart}
\vspace{-5mm}
\end{figure}

The proposed algorithm for finding the hardware parameters greatly improved the search performance compared to exhaustive search methods. The left graph in Figure \ref{fig:comparison_chart} depicts a decrease in the number of evaluation cases required by the proposed method compared to the exhaustive search. This reduction is consistent across all model sizes, highlighting our proposed approach's ability to target optimal configurations more directly and with fewer computations. Similarly, the right graph in Figure \ref{fig:comparison_chart} shows a substantial decrease in execution times for the proposed method.

\subsection{Effectiveness of CHOSEN Compiler}
\begin{figure}[tb]
\centering

\includegraphics[width=0.95\columnwidth]{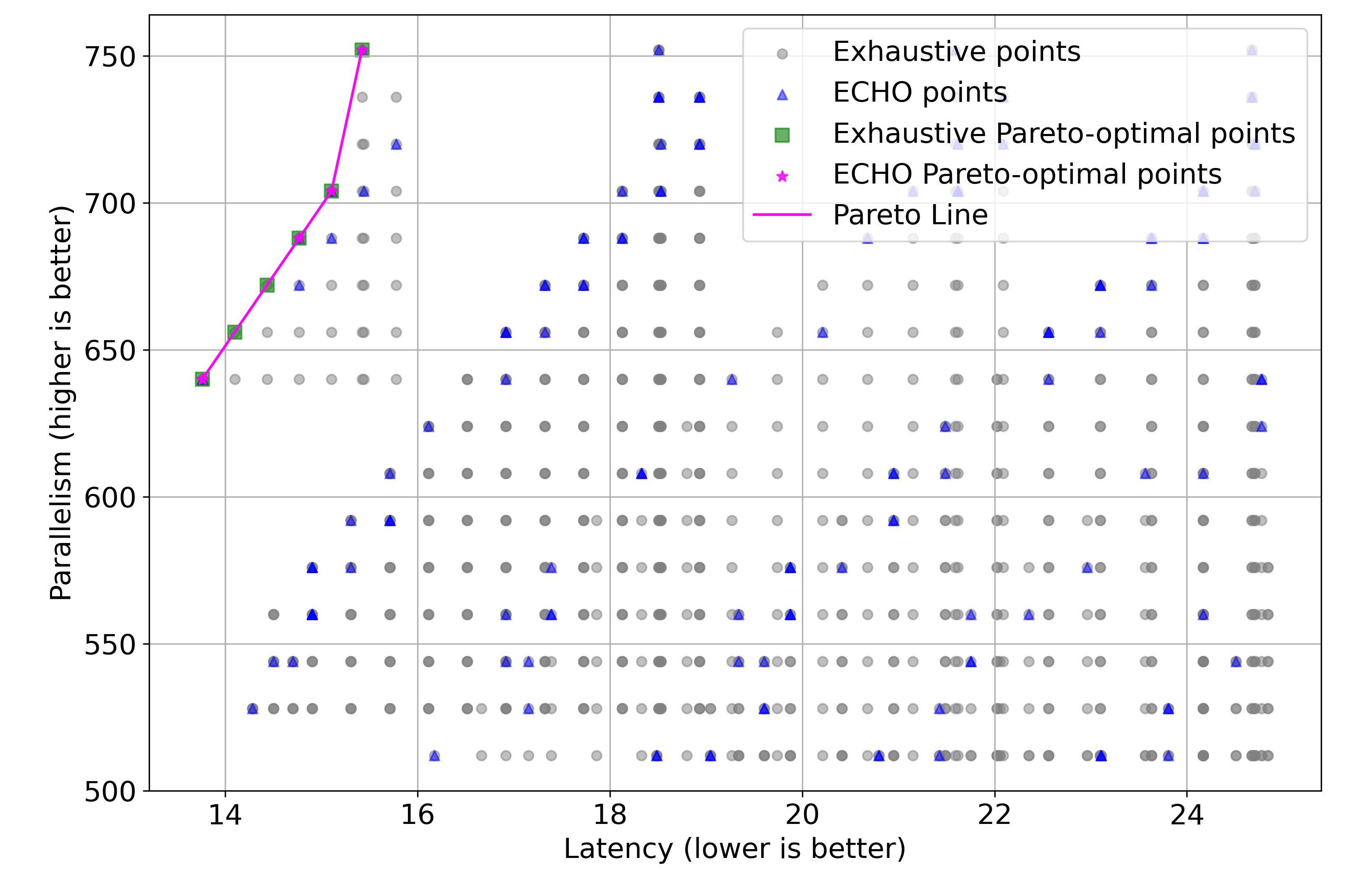}
\vspace{-4mm}
\caption{Pareto frontier comparison forf ViT Tiny model.}
\label{fig:pareto_tiny}
\vspace{-4mm}
\end{figure}

\begin{figure}[tb]
\centering

\includegraphics[width=0.95\columnwidth]{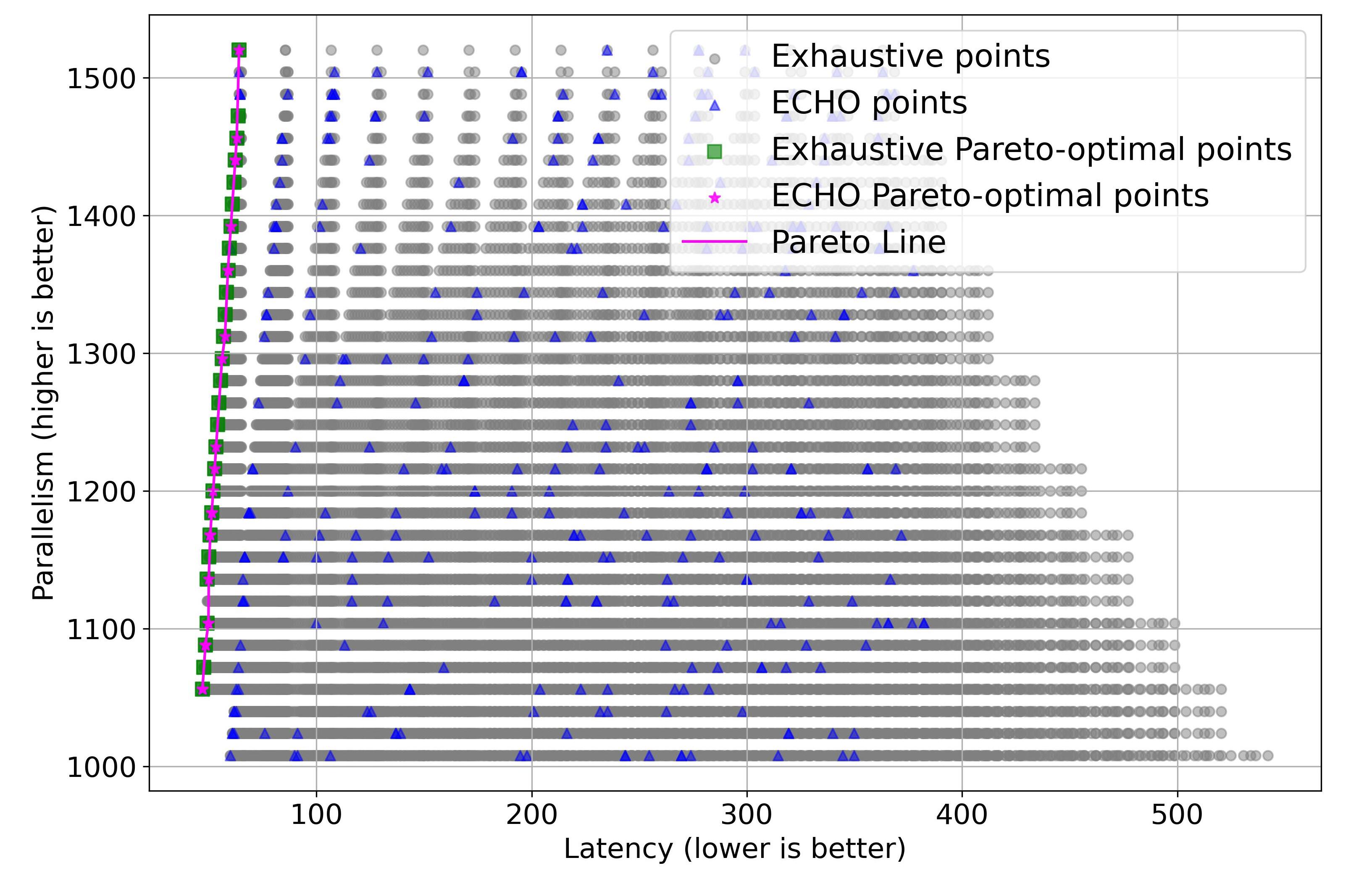}
\vspace{-4mm}
\caption{Pareto frontier comparison for ViT Small model.}
\label{fig:pareto_small}
\vspace{-4mm}
\end{figure}

In this section, we assessed the effectiveness of the CHOSEN's algorithm in finding the optimal parameters for the best performance of our inference engine. The proposed algorithm demonstrates substantial effectiveness in optimizing transformer models, particularly in its ability to identify Pareto optimal evaluation points with fewer evaluated points compared to exhaustive methods. This capability is illustrated in the plots for ViT Tiny and ViT Small models, as depicted in Figs. \ref{fig:pareto_tiny} and \ref{fig:pareto_small}.

Pareto frontier, represented by the magenta line in the plots, connects points that offer the best trade-off between latency and parallelism. The proposed approach captures nearly all the exhaustive search's Pareto-optimal points, demonstrating its efficiency. The proposed algorithm significantly reduces the number of evaluations, focusing only on configurations potentially leading to optimal outcomes. This targeted exploration is evident in the density of points along the Pareto frontier compared to the broader scatter of the exhaustive points. By focusing the search around the most promising areas of the solution space, the CHOSEN's algorithm ensures that the compiler's resources are not wasted on evaluating sub-optimal configurations.

\begin{table}[tb]
\centering
\caption{Hardware metrics for DeiT-B Implementation.}
\vspace{-3mm}
\label{tab:hardware}
\resizebox{\columnwidth}{!}{
\begin{tabular}{c c c c c c c}

\toprule
\textbf{HW Design}   &\textbf{HW Module} & \textbf{BRAM36}  & \textbf{URAM} & \textbf{DSP}  &  \textbf{LUT}& \textbf{FF} \\
\midrule[\heavyrulewidth]
\textbf{Auto-Vit-Acc\cite{DBLP:conf/fpl/LiSLMYX0LLWLF22}} & {-} & {-} & {-} & {2066} & {128K}& {-}  \\
\midrule[\heavyrulewidth]

\multirow{2}{*}{\textbf{ME-ViT \cite{DBLP:conf/hipc/MarinoZP23}}} & {ME-PE} & {288} & {-}   &{1024} & {192K}  & {137K} \\
{} & {Multi ME-PE} & {1440} & {-}   &{5120} & {960K}  &{685K} \\

\midrule[\heavyrulewidth]
\multirow{4}{*}{\textbf{CHOSEN-ViT}} & {1D PE array} & {180} & {192} & {848} & {101K}  & {73K} \\
{} & {Controller \& Scheduler} & {120} & {0}  &{9} &{11K} & {8K} \\
{} & {Non-linear functions} & {0} &{0}& {116} & 17K & 15K    \\
{} & {Total} & {1440} & {192}  & {6225}  &{ 833K} & {607K} \\
\bottomrule
\end{tabular} }
\vspace{-6mm}
\end{table}

\subsection{Hardware Cost}
\label{subsec:hard_cost}
Table \ref{tab:hardware} details the hardware metrics achieved by implementing CHOSEN-ViT. By utilizing the rapid and hardware-compatible approximations introduced by PEANO-ViT \cite{sadeghi2024peano}, the resource usage associated with hardware-intensive and costly iterative methods for exact non-linear implementation has been greatly diminished. Furthermore, Table \ref{tab:hardware} provides the resource utilization breakdown for each module in each computing kernel of CHOSEN-ViT. Please note that the hardware metrics for DeiT-Base implementation are reported. We used eight computing kernels in parallel. In processing non-linear functions such as normalization, softmax, and GELU, we simultaneously handle 16 elements, resulting in a Level of Parallelism (LoP) of 16. This LoP can be adjusted to align with resource availability and latency objectives, making CHOSEN a versatile framework for enhancing the speed of machine learning tasks. Increasing the LoP or the number of computing kernels enhances processing speed but may lead to higher resource consumption. As can be seen, our implementation can use higher DSP numbers for the matrix-matrix multiplication while having fewer LUTs and FFs due to the proposed approximations. 

\subsection{Performance Comparison with the State-of-the-Art ViT Accelerators}
\label{subsec:hard_cost}
We compare the performance obtained by CHOSEN on different ViT models with those of the prior work. To have meaningful and fair comparisons, we compare our results only with works that have used the same or a similar FPGA board as ours, with comparable resources. Table \ref{tab:performance} shows the performance comparison between CHOSEN-ViT and prior work references on ViT models on the ImageNet dataset. CHOSEN-ViT outperforms the state-of-the-art ViT accelerators by delivering 1.5x and 1.42x higher frame-per-second on DeiT-S and DeiT-B. CHOSEN-ViT executes at a frequency of 200 MHz. At the same time, ME-ViT \cite{DBLP:conf/hipc/MarinoZP23} reported a frequency of 300 MHz. ME-ViT \cite{DBLP:conf/hipc/MarinoZP23} delivers higher FPS in the case of DeiT-T as they can bring all the weights required for this model on-chip and they do not need to tile the memory for this model as it manually optimized. Instead, CHOSEN presented an automated framework that can be used for any transformer-based model that is not limited to ViTs.

\begin{table}[tb]
\centering
\caption{Performance Comparison \& CHOSEN-ViT Configuration.}
\vspace{-2mm}
\label{tab:performance}
\resizebox{\columnwidth}{!}{
\begin{tabular}{c c c c c c c}

\toprule
\textbf{ViT Model} & \textbf{Hardware Design}  & \textbf{ \( P_n\)} & \textbf{ \( P_m\)}  & \textbf{ \( T_n\)} & \textbf{ \( T_m\)}  &  \textbf{FPS} \\
\midrule[\heavyrulewidth]
\multirow{2}{*}{\textbf{DeiT-T}} & ME-ViT \cite{DBLP:conf/hipc/MarinoZP23} & {-} & {-} & {-} & {-} & {\textbf{298}} $^\dagger$ \\
{} & {\textbf{CHOSEN-ViT}} & {35} & {16} & {210} & {576}  &{171} \\
\midrule[\heavyrulewidth]
\multirow{2}{*}{\textbf{DeiT-S}} & ME-ViT \cite{DBLP:conf/hipc/MarinoZP23} & {-} & {-} & {-} & {-} & {88} \\
{} & \textbf{CHOSEN-ViT} & {99} & {16} & {198} & {1600}  &\textbf{132} \\
\midrule[\heavyrulewidth]
\multirow{3}{*}{\textbf{DeiT-B}} & ME-ViT \cite{DBLP:conf/hipc/MarinoZP23} & {-} & {-} & {-} & {-} & {21} \\
{} & Auto-ViT-Acc \cite{DBLP:conf/fpl/LiSLMYX0LLWLF22} & {-} & {-} & {-} & {-} & {26} \\
{} & \textbf{CHOSEN-ViT} & {102} & {16} & {212} & {3072}  &\textbf{37} \\
\bottomrule
\end{tabular} }
\begin{flushleft}
\hspace{5pt} \scriptsize $^\dagger$ They have on-chip memory storage for weights and have claimed 300 Mhz frequency.
\end{flushleft}
\vspace{-7mm}
\end{table}


\section{Conclusion} 
\label{sec:conc}
This paper introduces CHOSEN-ViT, a compiler-to-hardware optimization stack for deploying Vision Transformers (ViTs) on FPGAs. CHOSEN framework features a multi-kernel accelerator architecture that maximizes bandwidth by leveraging multiple DDR memory banks. The CHOSEN compiler enhances computing kernel performance and memory efficiency while managing data movements statically. Compared to leading ViT accelerators, CHOSEN-ViT delivers throughput improvements of 1.5x for DeiT-S and 1.42x for DeiT-B models.

\bibliography{CHOSEN}

\end{document}